\documentclass{article}

    \PassOptionsToPackage{numbers, compress}{natbib}
\usepackage[final]{neurips_2019}

\usepackage[utf8]{inputenc} % allow utf-8 input
\usepackage[T1]{fontenc}    % use 8-bit T1 fonts
\usepackage{hyperref}       % hyperlinks
\usepackage{url}            % simple URL typesetting
\usepackage{booktabs}       % professional-quality tables
\usepackage{amsfonts}       % blackboard math symbols
\usepackage{nicefrac}       % compact symbols for 1/2, etc.
\usepackage{microtype}      % microtypography
\usepackage{graphicx}
\usepackage{algorithm}
\usepackage{algorithmic}
\usepackage{array}
\usepackage{multirow}
\usepackage{subcaption}
\usepackage{wrapfig}
\usepackage{amsmath}
\usepackage{amssymb}
\usepackage{amsthm}
\usepackage{mathtools}
\usepackage{enumitem}

\title{Tri-graph Information Propagation for Polypharmacy Side Effect Prediction}
\author{%
  Hao Xu$^1$, Shengqi Sang$^{2,3}$, and Haiping Lu$^1$  \\
  $^1$Department of Computer Science, University of Sheffield, UK\\
  $^2$Department of Physics and Astronomy, University of Waterloo, Canada\\
  $^3$Perimeter Institute for Theoretical Physics, Waterloo, Ontario, Canada\\
  \texttt{hxu31@sheffield.ac.uk}, \texttt{s4sang@uwaterloo.ca}, \texttt{h.lu@sheffield.ac.uk } %\\
}

\begin{document}

\maketitle

\begin{abstract}
The use of drug combinations often leads to polypharmacy side effects (POSE). A recent method formulates POSE prediction as a link prediction problem on a graph of drugs and proteins, and solves it with Graph Convolutional Networks (GCNs). However, due to the complex relationships in POSE, this method has high computational cost and memory demand. This paper proposes a flexible Tri-graph Information Propagation (TIP) model that operates on three subgraphs to learn representations progressively by propagation from protein-protein graph to drug-drug graph via protein-drug graph. Experiments show that TIP improves accuracy by 7\%+, time efficiency by 83$\times$, and space efficiency by 3$\times$. 

% We propose Tre-graph Interaction Propagation (TIP) network to predict POSE more efficiently and accurately. It solves the problems of low efficiency and high memory consumption in the existing graph convolutional approaches for POSE modeling.
\end{abstract}

\section{Introduction}
\label{intro}
% Patients may suﬀer from side eﬀects caused by drug interactions when they are taking many drugs at the same time.  Recent works with graph convolution neural networks have successfully predicted thousands of possible side eﬀects between hundreds of side eﬀects \cite{decagon}. However, these methods are faced with the problems of low training eﬃciency and taking up a lot of memory.

% Model We propose TIP, a novel model to extract relationship information from protein-protein interaction (P-P) graph, propagate information from P-P graph to drug-drug multi-relational interaction (D-D) graph via protein-drug interaction (D-P) graph and predict relationships between nodes in D-D graph.

% Result By using this architecture of information propagation between graphs, the TIP model improves the information propagation eﬃciency of multi-relational convolution neural network and reduce its memory consumption signiﬁcantly. TIP model achieves the state-of-the-art performance in the task of predicting polypharmacy drug side eﬀects. The TIP model improves the accuracy of the baseline model by 7.2\%, reduces training time by at least 98\%, and reduces memory consumption by at least 66\%.

When treating complex or simultaneous diseases, patients often have to take more than one drugs concurrently, called \emph{polypharmacy}. This often causes additional side effects, i.e., \emph{polypharmacy side effects} (POSE) due to interactions between drugs. Graph convolutional network (GCN) is an emerging approach for graph representation learning \cite{repre, gae, gcn}. GCN-based drug representation learning has shown improved performance in POSE prediction \cite{multiview, kgc, deepddi, gamenet, decagon}. 

% GCNs \cite{decagon} has been used  to predict thousands of side effects between hundreds of drugs, by constructing a \emph{multi-modal graph} which contains \emph{POSE clinical records} (i.e. drug-drug interactions (P-P) with side effects as labels) and \emph{pharmacological information} (i.e. protein-drug interactions (P-D) and protein-protein interactions (P-P)), as shown in Figure.\ref{fig:graph}. 
\begin{wrapfigure}{r}{0.45\textwidth}
\vspace{-5mm}
  \hfill
%   \fbox{\rule[-.5cm]{0cm}{4cm} \rule[-.4cm]{3cm}{0cm}}
  \includegraphics[width=0.5\columnwidth]{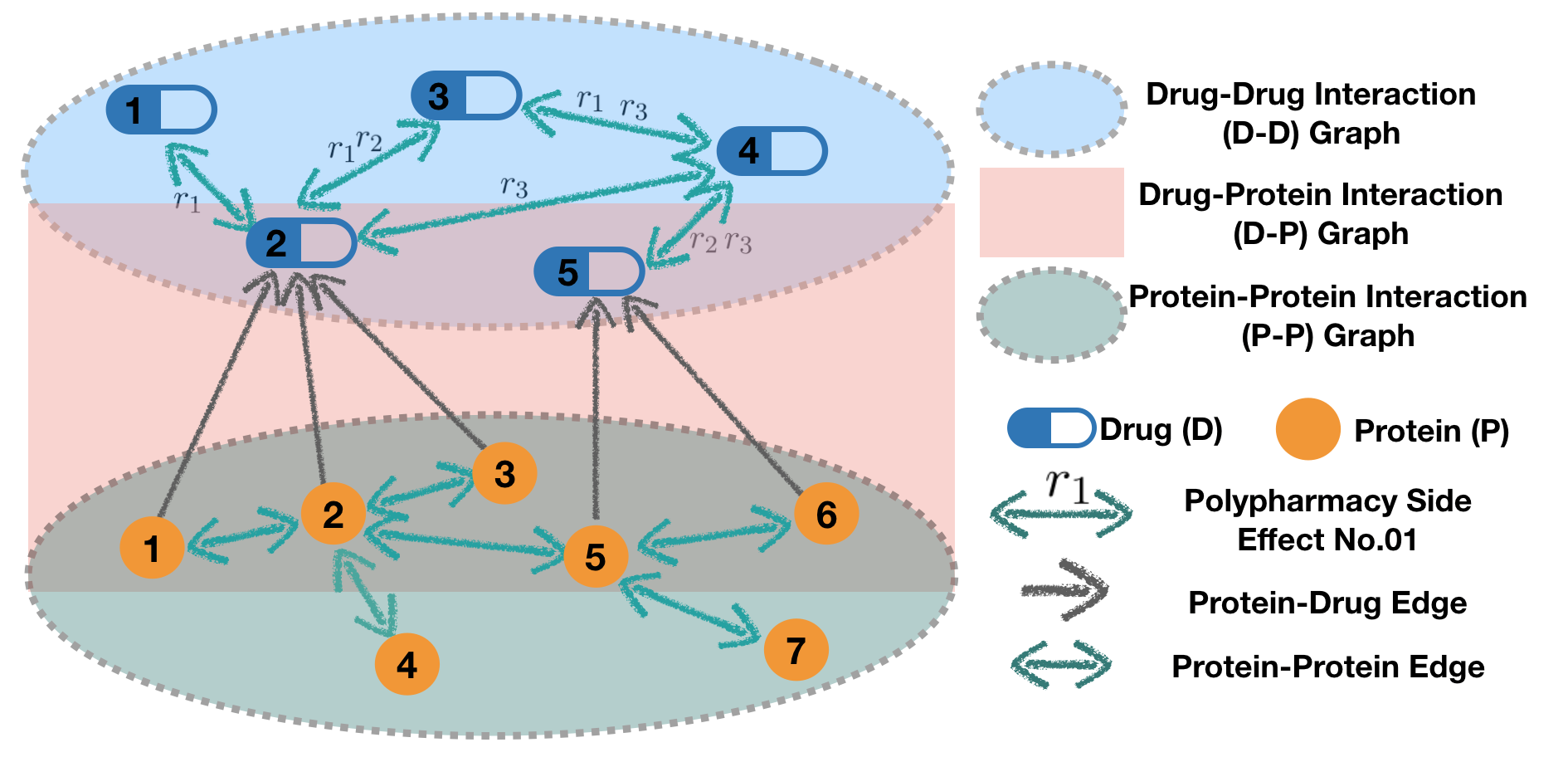}
  \caption{A multi-modal biomedical graph with two types of nodes: Drug (D) and Protein (P), and three types of edges: Protein-Protein (P-P) edges labeled with $b$ (fixed), Protein-Drug (P-D) edges labeled with $t$ (fixed), and Drug-Drug (D-D) edges labeled by a side effect $r\in R$.}
  \label{fig:graph}
  %\vspace{-17mm}
  \vspace{-5mm}
\end{wrapfigure}
POSE prediction can be viewed as a link prediction problem. As shown in Fig. \ref{fig:graph}, a \emph{multi-modal graph} can be constructed using 1) drug-drug interactions (D-D) with side effects as edge labels, e.g., from \emph{POSE clinical records}, 2) protein-drug interactions (P-D) with edges labeled as $t$,  and 3) protein-protein interactions (P-P)) with edges labeled as $b$, e.g., from \emph{pharmacological information}. On such a graph, Zitnik et al. \cite{decagon} proposed a GCN-based \emph{Decagon} model to learn drug/protein representation via weighted aggregation of local neighbourhood information, with different weights assigned to different edge labels. It predicts all relationships between all nodes (drug/protein). This formulation enables the prediction of side effects that have strong molecular origins. However, due to the large number of nodes and possible edge labels, the aggregation operation has both high computational cost and high memory demand. 

% when propagating information to/from both drug and protein nodes and predicting all types of relationships.

Inspired by the Decagon model and motivated by its limitations, we propose a Tri-graph Information Propagation (TIP) model for improving prediction accuracy, and time and space efficiency, as shown in Fig. \ref{fig:encoder}. We start from the same multi-modal biomedical graph as in \cite{decagon}, constructed from three open BioSNAP-Decagon datasets \cite{snap}, as detailed in Table \ref{tab:dataset}. Instead of viewing the graph as a whole, we propose to view it as three subgraphs: the P-P graph, P-D graph and D-D graph, as in Figs.\ref{fig:graph} and \ref{fig:encoder}. TIP focuses on predicting relationships (side effects) in the D-D graph only rather than all relationships in the whole graph in Decagon. Thus, we treat drug nodes and protein nodes differently. Specifically, TIP has four steps: \textbf{1)} learn protein embedding on the P-P graph; \textbf{2)} propagate such embedding to the D-D graph via the P-D graph; \textbf{3)} learn the final drug embedding; \textbf{4)} predict the side effects on the D-D graph. 

TIP embeds proteins and drugs into different spaces of possibly different dimensions, rather than the same space and dimensions as in Decagon. This enables the propagation of flexible protein embedding to drug embedding as supplementary information. 
%proteins to drugs as learning a higher-level embedding of proteins. 
This brings three key benefits: \textbf{1) Flexibility}. We design three information propagation GCN modules corresponding to the first three TIP steps and two ways to combine protein and drug information in the P-D graph (step 2). Thus, we have the flexibility to set the number of GCN layers to control the order of neighborhood considered in each module.
\textbf{2) Efficiency}. Separate embedding of proteins and drugs can greatly improve the time (83$\times$) and space (3$\times$) efficiency of GCN-based representation learning and information propagation for them,  
%leading to significant 98\%, with at least 66\% memory consumption reduction. 
\textbf{3) Accuracy}. More focused learning of drug representation makes better use of available data sources and can lead to improved POSE prediction, e.g., by 7.2\% in our experiments.

% Specifically, the protein embedding/vectors are learned in the D-D graph and propagated to the protein-drug interaction (P-D) graph; the learned protein embedding and drug information then be integrated and propagated to D-D graph on which final drug representation learning and POSE prediction are done. 

% The intuition behind the model design is when predicting drug-drug relationships, we do not need to propagate information between all the nodes and predict all the relationships between nodes in the whole graph. For example, as shown in Table \ref{tab:dataset}, there are 19081 proteins but only 645 drugs, and the number of protein interactions is 14k times the average number of drug interactions per side effects. When passing information between proteins and drugs by relation types, it needs too many operations, which causes inefficiency issue and huge memory consumption. 
\par

\section{Tri-graph Information Propagation (TIP)}
\label{model}
% As shown in Figure.\ref{fig:graph}, we construct the same multi-modal biomedical graph as \cite{decagon} did, using three opened BioSNAP-Decagon datasets \cite{snap}. See details in Table \ref{tab:dataset}. We view the graph as three subgraphs and design three modules correspondingly for drug embedding learning (See Figure.\ref{fig:encoder}). We consider the POSE prediction task as a graph completion problem which aims to find the undiscovered edges and labels on the D-D graph.
% \subsection{Dataset and Problem Formulation}
% \textbf{Dataset and Problem Formulation}
% As shown in Figure.\ref{fig:graph}, we construct the same multi-modal biomedical graph with protein node (P) and drug node (D) for polypharmacy side effect modeling as \cite{decagon} did, using opened BioSNAP-Decagon dataset \cite{snap}. See details in Table \ref{tab:dataset}.

% We here consider POSEP task as a graph completion problem which aims to find the undiscovered edges and labels on the graph. Specifically, we extract the representation of the drugs from the defined graph $G$, and predict the probability of all possible side effects of a queried drug pair. 

% opened three datasets from Stanford Biomedical Network Dataset Collection (SNAP) \cite{snap}. These three SNAP-Decagon datasets \cite{decagon} is used to construct three subgraph: protein-protein interaction (P-P) graph, protein-drug interaction (P-D) graph and drug-drug interaction (D-D) graph. 
\begin{figure}[t]
	\centering
	\includegraphics[width=\columnwidth]{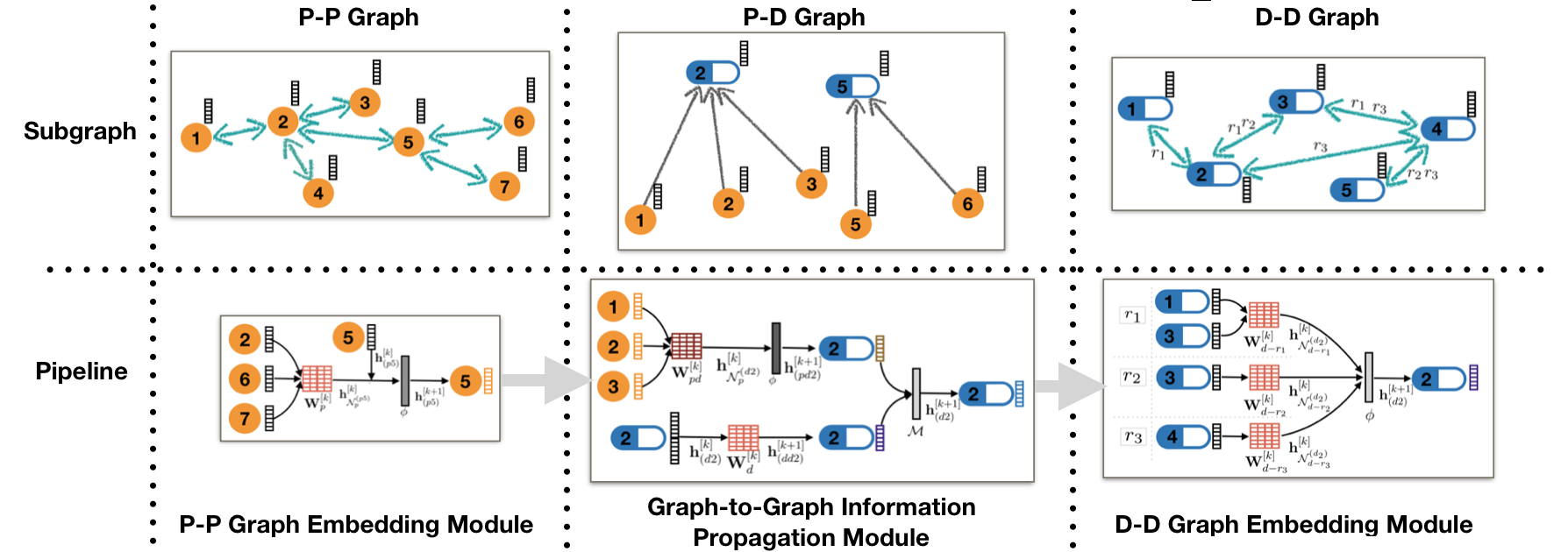}
	\caption{Information propagation in TIP encoder.}
	\vspace{-4mm}
	\label{fig:encoder}
\end{figure}

\begin{table}[t]
  \caption{BioSNAP-Decagon \cite{snap} datasets. (P) denotes protein node, and (D) denotes drug node.}
  \label{tab:dataset}
  \centering
  \begin{tabular}{l c r r c}
    \toprule
    \textbf{Dataset}     & \textbf{Nodes}     & \textbf{Edges} & \textbf{Unique Labels} & \textbf{Graph Name}\\
    \midrule
    PP-Decagon & 19081(P) & 715612 & 1 & P-P graph\\
    GhG-TargetDecagon & 3648(P), 284(D) & 18690 & 1 & P-D graph\\
    ChChSe-Decagon & 645(D) & 63473 & 1317 & D-D graph\\
    \bottomrule
  \end{tabular}
  \vspace{-5mm}
\end{table}

% \subsection{The Tri-graph Interaction Propagation (TIP) model}
% \textbf{The Tri-graph Interaction Propagation (TIP) model}
TIP follows the popular encoder-decoder framework \cite{repre}. Figure \ref{fig:encoder} shows the structure of the TIP encoder, within which Pharmacological information is propagated from P-P to D-D graph via P-D graph. The drug representation is produced by combining protein embedding and other available drug information. Further, drug embedding is used as input to the decoder to compute a set of side-effect-specified scores. Given a side effect and a drug pair, a higher score means the side effect is more likely to exist.% caused by this drug combination.
% We propose a non-linear, tri-graph information propagation neural network with an encoder-decoder perspective. As shown in Figure \ref{fig:encoder}, the information is propagated from P-P graph to D-D graph via P-D graph. The drug representations $H_d$ are yielded by the combination of protein embedding $H_p$ and drug information (i.e. $E_d$). The drug representations $H_d$ are than used in the decoder to compute a set of side-effect-specified scores $\{ g_{r_k} | r_k \in R\}$. Given a side effect and a drug pair, a higher score means the side effect is more likely caused by this drug combination.

% \subsection{TIP Encoder}
\textbf{TIP Encoder:} 
We follow the same \emph{Message Passing Neural Networks} (MPNN) framework \cite{message} as GCN \cite{gcn}, Decagon  \cite{decagon} and R-GCNs \cite{rgcn}. Our Encoder can be considered as a sequence of different MPNN cases. The protein and drug input features are $\textbf{V}_p \in \mathbb{R}^{N^p \times N^p}$ and $\textbf{V}_d \in \mathbb{R}^{N^d \times N^d}$, where $N^{p/d}$ is the total number of proteins/drugs. 

% Each of them is an identity matrix. 
% We employ two existing MPNN encoder \cite{gae, rgcn} to learn node embedding/vectors on P-P graph and D-D graph, and design two propagation operations on P-D graph passing information from P-P graph to D-D graph. 
\textbf{1) P-P Graph Embedding Module (PPM):} 
PPM is a GCN module \cite{gcn} used to learn protein embedding. The input of PPM module is the protein features $h^{0}=\textbf{V}_p $. The relation between two hidden layers is given by
\begin{equation}
	h_{(p_i)}^{k+1} = \text{ReLU} ( \frac{1}{c_i} \sum_{j \in \mathcal{N}_i}   W_p^k h_{(p_j)}^k+ h_{(p_i)}^{k} ),
\end{equation}
where $c_i = |\mathcal{N}_i|$ and $i$ is associated with a protein node $p_i \in P$. 

\textbf{2) Graph-to-Graph Information Propagation Module (GGM):} This module takes $\textbf{V}_d$ and the protein embedding generated by PPM to learn the embedding of pharmacological information associated with each drugs. It contains two units: \\
\textit{\textbf{2a)} Graph-to-Graph unit}: a one-layer MPNN with
\begin{equation}
h_{(d_i)}^H = \text{ReLU} ( \dfrac{1}{c_i} \sum_{j \in \mathcal{N}_i}  W_{h} h_{(p_j)} ),
\end{equation}
where $h_{(d_i)}^H$ can be regraded as a higher level representation of a subset of proteins, inspired by the subgraph embedding algorithm \cite{subgraph} which simply sums over the feature vectors of the involved nodes.\\ \textit{\textbf{2b)} Drug feature dimension reduction unit}: A linear transformation followed by an activation function:
\begin{equation}
h_{(d_i)}^D = \text{ReLU} ( W_{d} v_{(d_i)}).
\end{equation}
The output of GGM $h_{(d_i)}^{k+1}$ is the concatenation (\textbf{TIP-cat}) or the sum (\textbf{TIP-sum}) of $h_{(d_i)}^H$ and $h_{(d_i)}^D$.

\textbf{3) D-D Graph Embedding Module (DDM):}
This module is a R-GCN encoder with a basis-decomposition  regularization \cite{rgcn}. The update rule between layers is:
\begin{equation}
	h_{(d_i)}^{k+1} = \text{ReLU} \left(\sum_{r\in R}  \sum_{j \in \mathcal{N}_{r}^{i}}  \dfrac{1}{c_{i, r}} W_{r}^k h_{(d_j)}^k + W_o^k h_{d_i}^{k} \right) \quad W_r^{k} = \sum_{b\in[B]} a_{rb}^{k} V_b^{k},
\end{equation}
where $c_{i, r} = |\mathcal{N}_{r}^{d_i}|$ and $h^0 = [h^H, h^D] \text{ or } h^H + h^D$.  The weight $W_r^k$ was regularized by basis-decomposition \cite{rgcn}, which decomposes the matrix into the linear combination of a small number of basis matrices $V_b^{k} \in \mathbb{R}^{d^{l+1} \times d^{l}}$ with side-effect-specified coefficients $a_{rb}^{k}$.
% \begin{equation}
% 	W_r^{k} = \sum^B_{b=1} a_{rb}^{k} V_b^{k}
% \end{equation}
	
\textbf{TIP Decoder:}
% The TIP takes final drug representations $\mathbf{Z}_d$ learned from  TIP encoder, and computes the probability of side effect $r\in R$ given a pair of drugs embedding $(\mathbf{z}_i, \mathbf{z}_j)$, i.e. $p^{i, j}_r$. We use the DistMult Factorization (DF) \cite{distmult} to compute a $N^d \times N^d \times N^r$ score matrix $G$, where each elements $g^{i, j}_r$ is the score corresponding to the probability $p^{i, j}_r$.
% \begin{equation}
% p^{i, j}_r = \sigma(g_r^{ij}) = \sigma(\mathbf{z}_i^T \mathbf{M}_r \mathbf{z}_j),
% \end{equation}
% where $\mathbf{M}_r$ is a trainable side-effect-specified diagonal matrix.
TIP takes the final drug representation $\mathbf{Z}_d$ learned from  TIP encoder, and computes the probability $p^{i, j}_r$ of side effect $r\in R$ given a pair of drugs embedding $(\mathbf{z}_i, \mathbf{z}_j)$. For the POSE task we only care about predicting edges and edge labels on the  D-D graph. We consider using the DistMult factorization \cite{distmult} or a 2-layer neural network multi-label classifier as the decoder.

\textbf{1) DistMult Factorization decoder  (DF):} For the DF decoder \cite{distmult}, we first compute a $N^d \times N^d \times N^r$ score tensor $G =\{g^{i, j}_r\} $, and then get the probability by acting the sigmoid function on it:
\begin{equation}
p^{ij}_r = \sigma(g_r^{ij}) = \sigma(\mathbf{z}_i^T \mathbf{M}_r \mathbf{z}_j),
\end{equation}
where $\mathbf{M}_r$ is a trainable diagonal matrix associated with the side effect $r$.

\textbf{2) Neural Network Decoder (NN):}
 NN-decoder is a multi-classifier with each side effects corresponded to a classifier. It takes the concatenation of drug pair's representations as input and embeds it into a lower-dimensional space in the first layer. For second layer it predicts the probability of all the possible side effects with the sigmoid function.  

We will compare the performance of two decoders in the following chapter.
%\footnote{Source code, trained models, testing results and model variation examples are available online.} \footnote{a package for accelerating the model training of MPNN-based algorithms on GPU}
\section{Experimental Results and Discussions}
\label{exp}
We implement TIP in PyTorch \cite{torch} with PyTorch-Geometric package \cite{pytorch}. The code is available at {\url{https://github.com/NYXFLOWER/TIP}}. Hyper-parameter setting, model training, optimization  and performance measurement details are in the supplementary material. 
% For each side effect, we use $80\%$ edges in D-D graph for model training and the remaining $20\%$ for testing. 
% We measure the performance using: 1) AUPRC: area under precision-recall curve, 2) AUROC: area under the receiver-operating characteristic, and 3) AP@k: average precision for the top $k$ predictions for each side effect.
% The computing cost (i.e. training time and peak GPU memory usage) is evaluated by using pytorch\_memlab package \footnote{\url{https://github.com/Stonesjtu/pytorch_memlab}}. 

\textbf{Models and Baselines} As shown in Table \ref{tab:xor}, we study two TIP model implementations TIP-cat and TIP-sum with concatenation or sum in GGM, and two degenerated TIP (dTIP) models dTIP\textsubscript{D} and dTIP\textsubscript{P} focusing on modelling drug or protein, respectively. We compare them with two recent POSE prediction models reporting state-of-the-art performance on the same dataset: Decagon \cite{decagon} and DistMult \cite{distmult} (reported by \cite{kgc}).
% \footnote{DistMult is reported as the best performance model on BioSNAP-Decagon dataset currently by \cite{kgc}.}
We also study R-GCN \cite{rgcn}, which shows good performance on standard datasets. These models are described in detail in the supplementary materials.

% All these two implementations use a two-layer PPM, a one-layer GGM and a two-layer DDM. The different is:TIP-cat uses concatenation operation in GGM , while TIP-sum uses sum operation in GGM. We also explore the different module combinations and compare TIP performance with the following recent state-of-the-art POSEP model on Decagon-dataset: 1).Decagon \cite{decagon}: a multi-layer convolutional graph neural network; 2) DistMult \cite{distmult}: reported as the best performance model currently by \cite{kgc}.

\begin{table}[t]
		\caption{Performance comparison on the SNAP-Decagon dataset. The best result is in bold for each evaluation metric. For Decagon, we quote the accuracy score in \cite{decagon} (marked with $^{*}$) and estimate the space and time cost from sub-set implementation (indicated by $^+$). Acronyms are described Secs. 2 and 3. \textbf{ARCT}: architecture; \textbf{Mem}: peak memory usage; \textbf{TpE}: computational time per epoch (including training and testing score computation).}
		\vspace{1mm}
		\begin{tabular}{lcrrrrr}
			\toprule
			\textbf{Model}& \textbf{ARCT} & \textbf{AUPRC} & \textbf{AUROC} & \textbf{AP@50} & \textbf{Mem(G)} & \textbf{TpE(s)} \\
			\midrule
			Decagon & & $^*$0.832 & $^*$0.872& $^*$0.803 & $>$$^+$28 & $>$$^+$9600  \\
			DistMult & DF         & 0.835 & 0.859 & 0.834 & 9.25 & 41 \\
			R-GCN &DDM-DF  & 0.882 & 0.908 & 0.883 & 10.49  & 82  \\
			\midrule
			dTIP\textsubscript{D} & DDM-NN & 0.791 & 0.847 & 0.792 & 9.49  &  118 \\
			dTIP\textsubscript{P} & PPM-GGM-NN  & 0.746 & 0.743 & 0.733 & \textbf{ 6.38} & \textbf{29}  \\
			TIP-cat & PPM-GGM-DDM-DF  & 0.889 & 0.913 & \textbf{0.890} & 9.47  & 116  \\
			TIP-sum & PPM-GGM-DDM-DF & \textbf{0.890} &\textbf{0.914} &\textbf{ 0.890} & 9.47  & 115   \\
			\bottomrule
		\end{tabular}
		\vspace{-5mm}
		\label{tab:xor}
	\end{table}

% It took DistMult model the shortest time to achieve results comparable to the Decagon model. 
\textbf{Performance comparison} 
TIP-cat and TIP-sum are the top two performers, outperforming Decagon by 7.2+\% in AUPRC and much more in AUROC and AP@50. Compared to Decagon, TIP-cat and TIP-sum reduce Decagon's computational time by at least 98.9\% and the peak GPU usage by at least 66.1\%. 
TIP models achieve good performance because of the efficient information propagation between graphs. Learning the embedding of proteins in the P-P graph is efficient as all the propagation operations share the same trainable parameter at each layer. 
% The peak memory usage of both TIP-cat and TIP-sum are the same as expected. After integrating information between learned protein embedding and drug features, the dimension of the output of GGM is same. 
The most time and memory consuming part is the drug embedding learning on D-D graph, which takes $\sim 74\%$ of the total training time and hits the peak GPU memory usage of 9.47G. 

\textbf{Learning drug embedding with pharmacological information} Pharmacological information does contain drug-drug interaction information. By using it directly in dTIP\textsubscript{P}, we can get decent result with the shortest time. However, compared with R-GCN, additional pharmacological information in TIP-sum only improves the performance slightly.  In addition, the comparable performance of TIP-cat and TIP-sum has an interesting implication:
%implies the important conclusion:
% information propagation from protein to drug can be considered as learning a higher-level representation of a subset of proteins. This type of higher-level representation captures the relationship between proteins, and between proteins and drugs.
information propagation from PPM to GGM can be considered as learning a higher-level representation of a subset of proteins, which captures the relationship between proteins, and between proteins and drugs.

\textbf{Drug representation learning on D-D graph} Compared with DistMult that uses the dimension-reduced drug features directly (DF), the additional use of DDM in R-GCN (i.e., DDM-DF) improves over DF only by 5.6\% (in AUPRC), and the further additional use of PPM and GGM in TIP-sum (i.e., PPM-GGM-DDM-DF) improves over DF only by 6.6\%. This is because when using DDM, the drug can learn from its local neighborhoods and capture the relationship information. While protein-protein interaction and protein-drug interaction are extracted as additional drug features when using PPM-GGM. When decoding the drug embedding, The DF decoder outperforms the NN decoder by 11.5\% in accuracy and 43.9\% in time cost. However, the DF decoder requires more memory than the NN one.

% Embedding drugs in the D-D graph is helpful. Compare with the DistMult which use the dimension-reduced drug features directly, the additional use of DDM improves 5.6\% its performance and that of PPM-GGM improves 6.6\%. This is because when using DDM, the drug learns from its local neighborhoods and capture the relationship information, and when using PPM-GGM protein-protein interaction and protein-drug interaction are extracted as additional drug features. 

% Predominantly pharmacological information is useful for the POSE prediction task. Compared with DDM-DF, additional pharmacological information in TIP-sum improve the performance of DDM-DF. Even use the pharmacological information directly in PPM-GGM-NN,  we can get acceptable result with the shorted time.

% The decoder is also important for multiple-relationship extraction. When decoding the drug information, DistMult performance better than a simple neural network by  11.5\% with 43.9\% less time cost. However,  DistMult factorization requires more memory than NN, although the implementation of DF has used sparse tensor variables.

\textbf{Prediction of molecular-original side effects} We list side effects with 20 best and worst performance in TIP-cat in AUPRC score in Figs. 4 and 5 of the supplementary materials, which show consistent conclusion that TIP is particularly good at modeling side  effects with inter-molecular origins.  However, by comparing these side effects, we find that even if the model does not have access to pharmacological information, it can predict the side effects with molecular origins very well. As shown in Table \ref{tab:xor}, the R-GCN model with architecture DDM-DF achieves performance that is competitive with TIP-cat or TIP-sum. 
% This might be due to the imbalanced distribution of side effects in the dataset. The side effects caused by few drug combinations may more easy to predict.

\section{Conclusion}
\label{furture}
In this work, we proposed a new Tri-graph Information Propagation (TIP) model for predicting more than one thousand side effects between hundreds of drugs, using pharmacological information and drug-drug interaction clinical records. TIP has achieved state-of-the-art performance on POSE prediction task with much less training time and memory consumption. It can be further improved by using general optimization strategies. It can also be applied to other problems such as cancer risk or drug response prediction.% in systems biology. %bio-system modeling.

% \subsubsection*{Acknowledgments}

% \section*{References}
\medskip
\small

\bibliographystyle{plain}
\bibliography{neurips_2019}
%\end{document}
\newpage
% \begin{appendices}
% \renewcommand*{\theHsection}{\theHchapter.\the\value{section}}
% \newcounter{section}
\textbf{\Large Supplementary Materials}

This is the supplementary material, including detailed problem formulation, notation, information propagation between nodes and graphs in TIP, model variants definitions, experimental setup and results.

\section{Problem Formulation and Notation}

As shown in Figure.1, we construct a large multi-modal biomedical graph with Drug (D) nodes and Protein (P) nodes for polypharmacy side effect modeling. Given a set of drugs $V_d = \{d_i\}_{i \in [N^d]}$, a set of proteins $V_p = \{p_i\}_{i \in [N^p]}$ and a set of side effects $R =  \{r_i\}_{i \in [N^r]}$, where $N^{d/p/r}$ is the total number of drugs/proteins/side effects, the graph can be denoted as $G = \{V, E\}$, where $V = \{v_i | v_i \in V_d \cup V_p\}$.

In the graph $G$, edges are directed and labeled: $E = \{(v_i, q, v_i)\}$, $v_i, v_j \in V$ and $l$ is a label of edges ($q$ will be defined below). There are three edge types: protein-protein (P-P) edges , drug-protein (P-D) edges and drug-drug (D-D) edges, the labels of edges are associated with different edge types. Corresponding to the edge types, there are three subgraphs:

\begin{enumerate}
	\item Undirected P-P graph: $G_{p} = \{V_p, E_p\}$. The edges $E_p = {(p_i, b, p)}$ are labeled with $b$, $p_i, p_j \in V_p$.
	
	\item Undirected D-D graph: $G_d = \{ V_d, E_d \}$, where $E_d = \{(d_i, r_k, d_j)\}$ means that a pair of drugs $(d_i, d_j)$ can cause multi-pharmacy side effect $r_k$.
	
	\item Directed D-P graph: $G_h = \{ V, E_h \}$, $E_h = {p_i, t, d_i}$ is a set of edges directed from a protein to a drug with edge label $t$.
\end{enumerate}

As shown above, the $G_p$/$G_h$ have the same label $b$ / $t$, but the label of $G_d$ is chosen from $R$, where each $r_i\in R$ represents a side effect. Note that between a pair of drugs there might be more than one links with different labels (a pair of drug might cause more than one side effects). Use $Q = \{q | q \in \{b, t\} \cup R\}$ represents all kinds of labels.

We here consider POSE prediction task as a graph completion problem which aims to find the undiscovered edges and labels on the graph. Specifically, we extract the representation of the drugs from the defined graph $G$ i.e. $H_d = \{h_{d_i} | d_i \in D\}$, and predict the probability of all possible side effects of a queried drug pair $(d_i, d_j)$, i.e. $\{P_{r_k}(d_i, d_j) |r_k \in R \}$.

\section{TIP Encoder Design - An MPNN Framework Perspective}
In our TIP encoder, each module corresponding to a special case of the Message Passing Neural Networks (MPNN) framework \cite{message} on a graph. A simple differentiable MPNN framework on a graph $G'=\{V', E'\}$ is:
	\begin{equation}
	h_i^{(l+1)} = \sigma ( \sum_{m \in M_i} g_m (h_i^{(l)}, \mathcal{N}_i) ),
	\end{equation}
where $i$ is associated with a node $v_i \in V'$. The input of the framework $h^{(0)}$ is a node feature vector, and $h_i^{(l)} \in \mathbb{R}^{d^{(l)}}$ is the hidden state of this node in the $l^{th}$ layer of the neural network. $\mathcal{M}_i$ is the set of type-specified message passed in the form of $g_m(\cdot, \cdot)$ related to node $v_i$, and $g_m(\cdot, \cdot)$ is typically a neural network-like function of the node state $v_i$ and its neighborhood $\mathcal{N}_i$. 

\begin{figure}[ht]
	\centering
	\includegraphics[width=0.8\columnwidth]{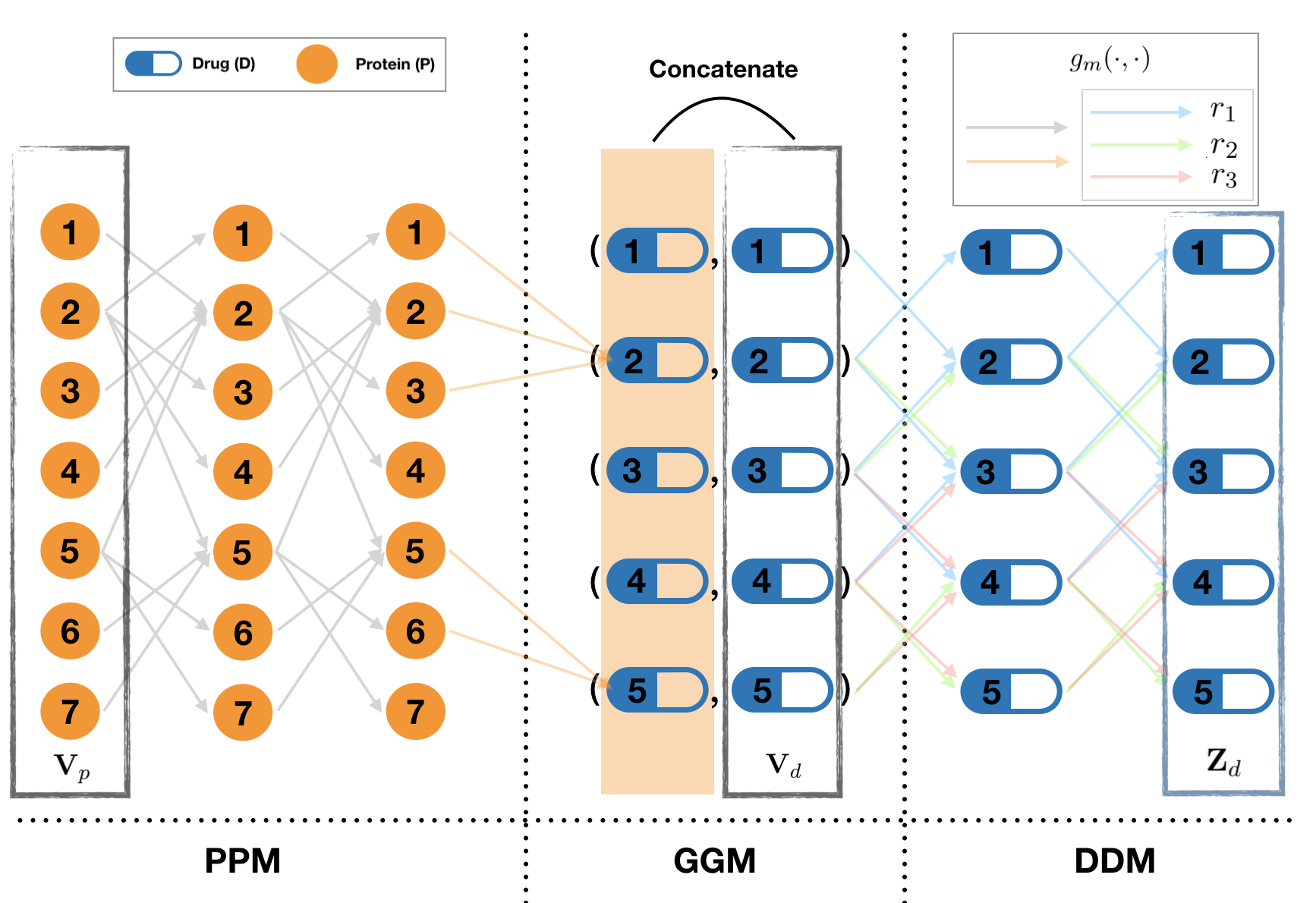}
	\caption{An example of information propagation between nodes and graphs in a TIP-cat implementation with a 2-layer PPM, a GGM with concatenation operation and a 2-layer DDM. }
	\label{fig:pass}
\end{figure}

Inspired by this architecture, we define the  tri-graph interaction propagation (TIP) encoder for calculating the update in each graphs forwardly. Figure.\ref{fig:pass} shows an example for information propagation between nodes and graphs in a TIP-cat implementation.

\section{Detail of Models}

The number of layers for PPM, GGM and DDM are set to (2, 1, 2) in all the experiments.

\paragraph{TIP-cat and TIP-sum} They both use a two-layer PPM with $d_{p}^1 = 32$ and $d_{p}^2=16$, a one-layer GGM and a two-layer DDM with $d_d^1=32$, $d_d^2=16$ and base number $d_d^b=16$. Their difference lies in the choice of aggregation function in GGM: TIP-cat uses concatenation with $d_{g}^p=16, d_g^d=48$, while TIP-sum uses summation with $d_g^p=d_g^d=64$. 

\paragraph{R-GCN} It's composed of a two-layer DDM with $d_d^1=32$, $d_d^2=16$ and a DistMult Factorization (DF) decoder. It models the D-D graph directly and is a special case of generic R-GCN for multi-relational link prediction \cite{rgcn}.

\paragraph{dTIP\textsubscript{D}} It uses the same DDM as DR-DF, and does not use any protein information. Drug embeddings are learned from DR module, and a 2-layer  neural network multi-classifier with $d_n^1=16$, $d_n^2=964$ is used as a decoder.

\paragraph{dTIP\textsubscript{P}} This variant uses the protein information and relationship information between drugs and proteins only to predict drug side effects. It uses a two-layer PPM with $d_{p}^1 = 32$ and $d_{p}^2=16$, a one-layer GGM $d_{g}^p=16, d_g^d=48$ with concatenation, and the same two layer NN decoder as DDM-NN.

\section{Experimental Setup}

\paragraph{Loss Function and Negative Sampling} We use cross-entropy loss to optimize model, aiming to assign higher probabilities to observed edges and lower probabilities to undiscovered ones. Given a set of positive samples $E_p' = \{(d_i, r, d_j) | r\in R\}$ , the negative samples $E'_n$ are sampled randomly from $R$ until $E'n \cap E'p = \emptyset$ \cite{gcn}.

\paragraph{Training and Testing data} We pre-processed the whole dataset (See \ref{tab:dataset}) by removing the side effects with less than 500 occurrence in the dataset.\footnote{It's the same pre-processing as in Zitnik et al. \cite{decagon}} For each side effect, we use $80\%$ of the total edges in D-D graph for model training and the remaining $20\%$ for testing.

\paragraph{Optimization} We use the Adam optimizer \cite{adam} with learning rate of 0.01 and train for 100 epochs for all the experiments. The TIP model is optimized end-to-end which means all trainable parameters in both encoder and decoder are trained together. Due to the Graph-to-Graph information propagation architecture of TIP model, the memory cost is much less than Decagon model \cite{decagon}. TIP model therefore is optimized by full-batch, which means the whole dataset is fed into the model in each epoch.

\paragraph{Model Implementation} We implement our TIP model  in PyTorch \cite{torch} with the PyTorch-Geometric package \cite{pytorch}. The evaluation of peak GPU memory usage uses the tools provided by pytorch\_memlab package\footnote{\url{https://github.com/Stonesjtu/pytorch_memlab}}.

\paragraph{Performance Measurement} We measure the performance using: 1) AUPRC: area under precision-recall curve, 2) AUROC: area under the receiver-operating characteristic, and 3) AP@k: average precision for the top $k$ predictions for each side effect. 4) The computing cost (i.e. training time and peak GPU memory usage) . 

\section{Prediction of Molecular-original Side Effects}
\begin{figure}[ht]
	\centering
	\includegraphics[width=\columnwidth]{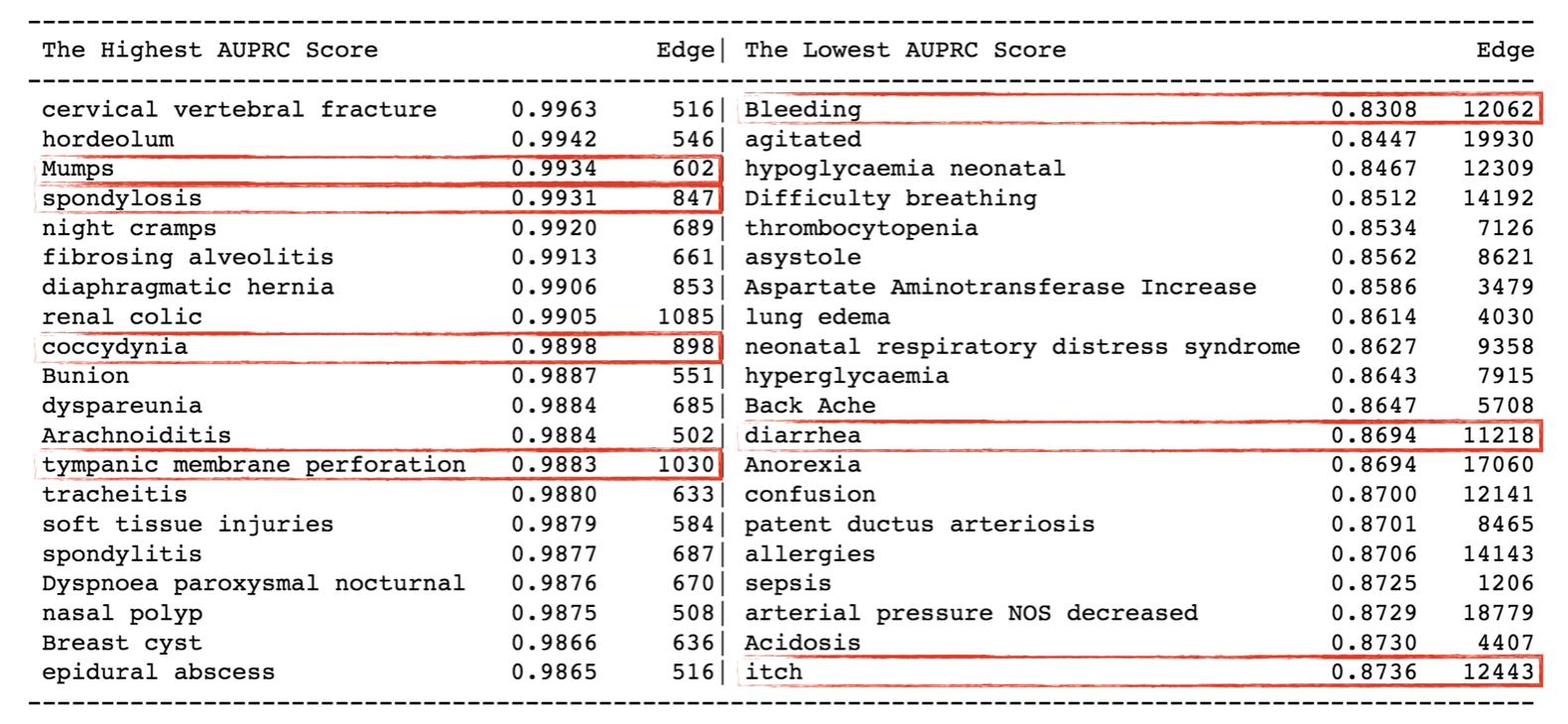}
	\caption{Side effects with the top 20 best and worst performance in TIP-cat on AUPRC scores. The side effects marked with red rectangular is in the side effect rank of the top 10 best/worst performance in \cite{decagon}}
	\label{fig:top20}
\end{figure}

\begin{figure}[ht]
	\centering
	\includegraphics[width=\columnwidth]{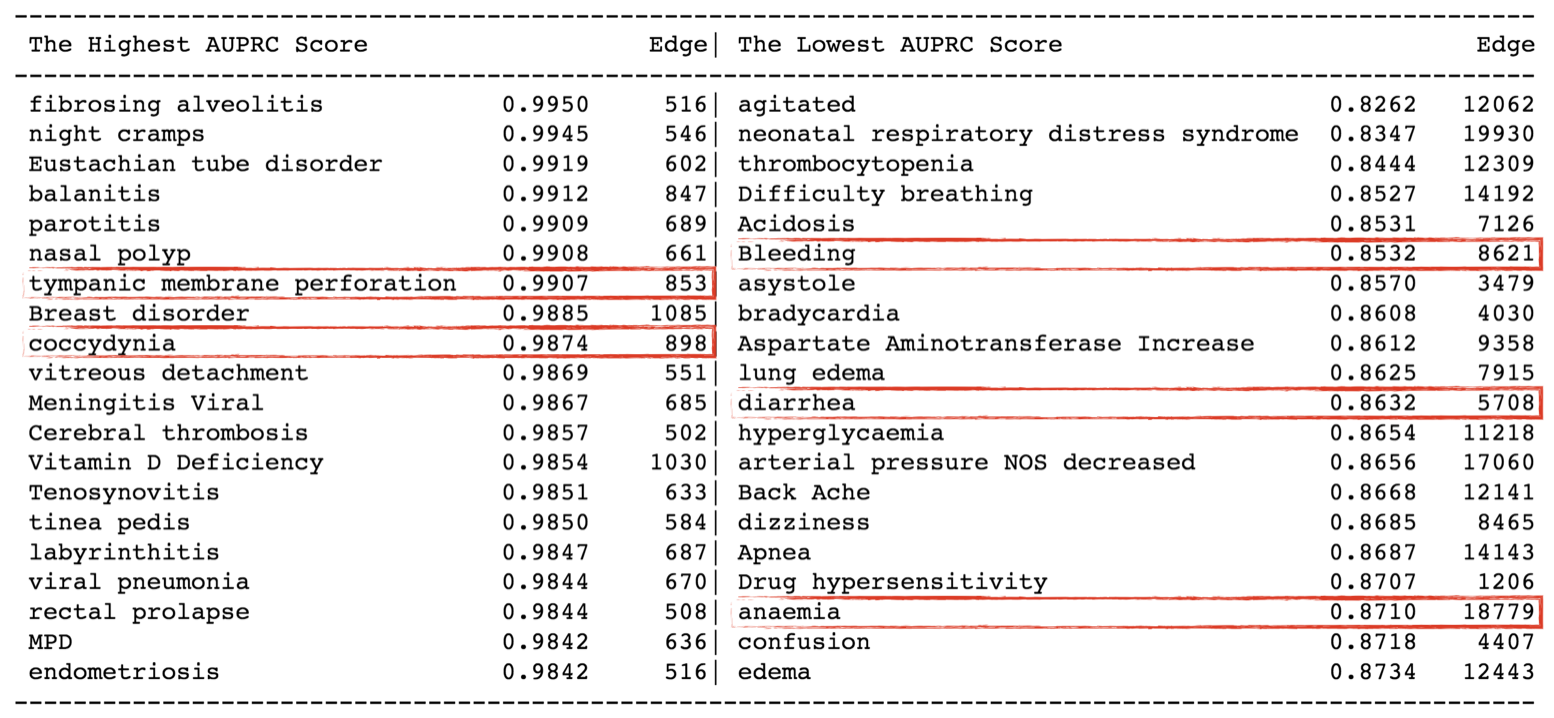}
	\caption{Side effects with the top 20 best and worst performance in DDM-DF model on AUPRC scores. The side effects marked with red rectangular is in the side effect rank of the top 10 best/worst performance in \cite{decagon}}
	\label{fig:top20-dd}
\end{figure}

We visualize the top 20 best and worst performance side effects in the DDM-DF model as shown in \ref{fig:top20} and \ref{fig:top20-dd}. Via comparing these figures, we find that even if model does not have pharmacological information, they can predict the side effects which have molecular origins very well. See more discussion in the main body.

% \end{appendices}

\end{document}